# Chrome Dino Run using Reinforcement Learning
## CS7IS2 Project (2019-2020)

Divyanshu Marwah, Sneha Srivastava, Anusha Gupta, Shruti Verma

marwahd@tcd.ie, ssrivast@tcd.ie, guptaa2@tcd.ie, sverma@tcd.ie

**Abstract.** Reinforcement Learning is one of the most advanced set of algorithms known to mankind which can compete in games and perform at par or even better than humans. In this paper we study most popular model free reinforcement learning algorithms along with convolutional neural network to train the agent for playing the game of Chrome Dino Run. We have used two of the popular temporal difference approaches namely Deep Q-Learning, and Expected SARSA and also implemented Double DQN model to train the agent and finally compare the scores with respect to the episodes and convergence of algorithms with respect to timesteps.

## 1 Introduction

Training an agent to take input from high dimensional space like vision and speech stands to be one of the long-standing challenges of reinforcement learning [1]. Recently introduced fusion of reinforcement learning and deep neural network provides a promising approach to resolve this class of problems. These algorithms can be used to train agents for playing arcade games and are known as Deep Q Network (DQN), which is based on off-policy reinforcement learning model Q Learning and uses a Convolutional Neural Network to learn the game-specific presentations. Games have always been a principal learning approach for AI and often used to illustrate major contributions in this field [2]. Following this tradition, we have chosen Chrome Dino Run for our implementation and comparison of algorithms.

Chrome Dino (also referred to as T-rex runner) is a game that appears in Google Chrome in offline mode. This can be accessed at URL, chrome://dino/. The enemies in this game are the obstacles in the form of cacti and birds. And the objective of the player/agent is to stay safe from these obstacles. The player runs on its own and the only state of possible actions is to jump or do nothing. The score increases for every step taken while the player is alive.

In chrome Dino, there is an indefinite number of states that are possible (since it is an infinite canvas of obstacles at different distances), and they are given to the model as an input in the form of convoluted 4x4 array of pixels. We aim to compare different temporal difference approaches for RL and see which performs best in this kind of environment. We have not provided any game-specific information to the network and let the model learn the next state based on the knowledge gained over the course of previous actions taken. The input given to the model is just the video signal (broken down into a set of images using a CNN), set of possible actions (jump or do nothing), reward and termination state (end game - the player died). Based on these given inputs the model learns when to take which action – similar to how a human plays the game. Also, we have kept the hyperparameters constant across all the algorithms to check the difference in convergence rates and compare their performances.

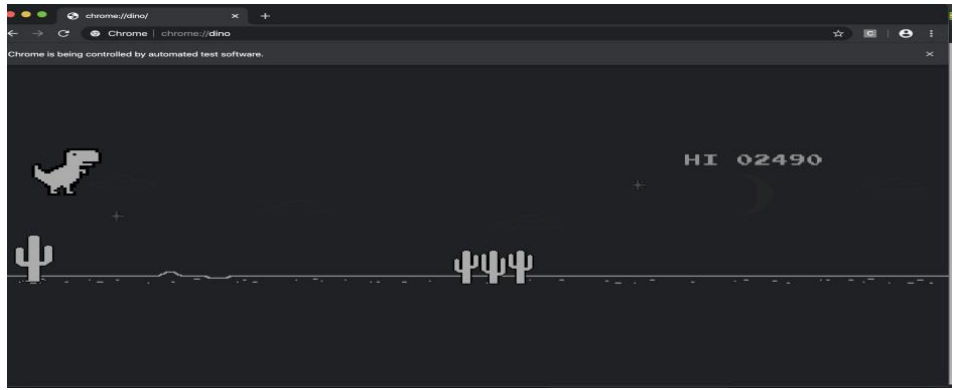

*Figure 1 Chrome Dino Run Game*

In this paper, we have implemented Deep Q-Network (DQN), Expected SARSA, and Double DQN algorithms to train an agent to play Chrome Dino run. By implementing these algorithms, we intend to answer the following research questions:
**RQ1:** Implementation and Comparison of on policy (Expected SARSA) with off policy (Deep Q Learning) learning models on training an agent to play this game.
**RQ2:** Implement different state of the art variants of Deep Q-Network models and compare their performances for score achieved in given timesteps of training.
By working on these research questions, we intend to develop in-depth knowledge in on-policy and off-policy reinforcement learning paradigms, development of convolutional neural networks, use of CNN along with the RL models and study the impact of variations in CNN model on the performance of our agent.

The paper is organized as follows: related work in this field is discussed in section 2, problem definition and different algorithms are described in detail in section 3. Section 4 presents the comparison, challenges, and discussion of the results achieved. And finally, the conclusion and future work are outlined in section 5.

## 2      Related Work

In recent years, there have been several pieces of research showcasing the implementation of AI models using reinforcement learning algorithms. [1] presented the novel approach of using deep neural network with RL to solve the complex problem of training agents for games by providing visual inputs. Many AI agents have been developed since then to play videogames such as Atari, Mario, etc. Exploring these researches, we decided to build an AI model for the Dino Run game using different RL algorithms and compare their performances with each other. Model-free based reinforcement learning is specifically quite powerful in learning the next states in such complex scenarios, that is, it learns the next action without using the probabilistic transition distribution [3]. Q-learning is one such approach which is an off-policy temporal difference control policy and finds the next action greedily [4].

Deep Q-Learning is a combination of the Q-Learning algorithm and deep neural networks. A research paper [1] demonstrates the implementation of the first DQN model in a video game. DQN is a method that has many advantages over Q-Learning such as it has greater data efficiency and reduces the variance of the updates [4]. Likely, Double DQN is an improved version of DQN. As shown in research, Q-Learning and DQN overestimate action values under certain conditions so Double DQN has been introduced which not only reduces the overestimation but also improves the performance on the game [5]. In DQN, the same values are used for selecting and evaluating an action whereas, in Double DQN, the selection and evaluation are decoupled i.e. two value functions are learned, one for estimating the value of the greedy policy and other for evaluating the value for this policy [5]. Another research presented Expected SARSA which is an on-policy reinforcement learning algorithm [6]. It is a variation of SARSA and we compare its performance with DQN to observe the comparison between on-policy and off-policy algorithms. It has been demonstrated in the paper that under the same conditions, expected SARSA performs better than SARSA and Q-Learning [6]. These algorithms work on the temporal-difference method which learns without having any prior knowledge of the environment and learns based on the experience replay mechanism. Q-learning updates incrementally and has a slow convergence rate when there is low variance in the next state, given the same pair of action and state. Experience replay alleviates this issue by iterating over the same data multiple times. It fits best with Dino run since there is an endless number of states with less variance in the next states and action and use of experience replay will help in faster convergence resulting in better training of the agent. Hence, we chose these three algorithms to verify and represent which algorithm outperforms the other two algorithms.

## 3    Problem Definition and Algorithm

The use of neural networks has given an opportunity to figure out solution of problems in complex and dynamic environments. Artificial Intelligence has moved from its humble start of object detection to its much complex applications like Google's driverless cars. Different reinforcement learning algorithms have their own pros and cons and the trade-off between optimal solutions and safe solutions is the key to finding the best algorithm for the given problem.

The goal of Reinforcement learning is to learn what to do to maximise the reward score in any situation.

Our problem statement of the Chrome Dino Run has the model learning to figure out a way to jump over obstacles like the cacti and avoid birds and run on a plain field based on the actions it takes and its subsequent rewards. The optimal Q-value that comes out of the neural network part of the model should be able to satisfy the Bellman Equation. The model uses trial and error to improve its performance and for this relies on Exploration to find out the best course of action and Exploitation to increase the reward depending on its learning. Here we use Epsilon-Greedy Algorithm to achieve the same. The inputs, output, use of hyperparameters remain the same across different algorithms; and the major difference is in the training of the agent for building our model. We evaluate the performance of different algorithms by comparing their convergence rates, maximum scores achieved in specific timeframe, and the average episode lengths.

a) The code uses selenium to make an interface between our python code and the browser-JavaScript code because the game is in JavaScript.
b) We then add a level of abstraction in the code by segregating the agent (dinosaur) which would take actions and the corresponding game states for controlling the agent, by creating 2 classes.
c) We have visual input for our algorithms. We grab 4 consecutive screenshots of the game frames by using selenium (as the game uses HTML Canvas) and then proceed to pre-process them to reduce their dimensionality.
d) We use OpenCV to pre-process the image. The stack of four images (for our input) which is of a large resolution (600*150*3) *4 is reduced to 80*80 and grey scaled. We use 3 convolutional layers (in CNN built using Keras and Tensorflow) to resize these images before flattening them to a dense layer. This forward feeds into the output layer which gives 2 values of Q as output, 1 for each action state (jump and do nothing). This architecture is shown in *Figure 2*.
e) The model is then trained with respect to no initial action and initial state (s_t), the agent observes for the OBSERVATION number of steps and it saves the experience it gains in the Replay Memory, a batch is trained from this memory and game gets restarted on the agent's death.

Under a given policy π, the true value of an action A in a state s, we make the standard assumption that future rewards are discounted by a factor of γ per time-step, and define the future discounted return at time t as:

$$R_t = \sum_{t'=t}^{T} \gamma^{t'-t} r_t$$

where T is the time-step at which the game terminates. [1]

### 3.1 DQN (Deep Q Networks)

The combination of deep neural networks with Q Learning is called as Deep Q Network Learning. Q learning might work well in small state space but with more complex and sophisticated environments it drastically reduces in performance. The environment in a video game will be quite large and the actions that can be taken are multiple with each state that can be represented as pixels. To iteratively store q values in such a large environment is computationally expensive. We make use of deep neural networks to estimate the q values in each state action pair. The optimal q value that comes out of the neural network part of the model should be able to satisfy the Bellman Equation.

The addition of Experience Replay enhances the performance of the DQNs [1]. It stores the states, actions, transitions, rewards and terminal states and makes batches to update the q values [1]. Four frames are grabbed and convoluted and taken as input.
Update Rule:

$$Q^*(s,a) = \mathbb{E}_{s'\sim\varepsilon}[r + \gamma \max_{a'} Q^*(s',a') \mid s,a]$$

It is an off-policy algorithm that focuses on finding the maximum q value and chooses the next step in a greedy manner.

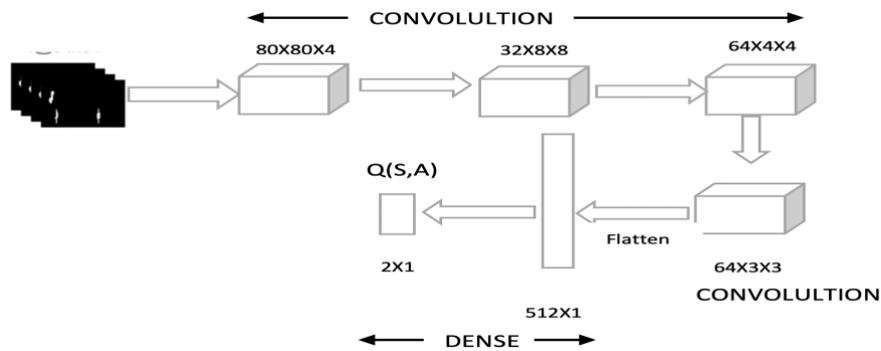

*Figure 2 CNN architecture for Deep Q Learning Model*

### 3.2 Expected SARSA

SARSA stands for State-Action-Reward-State-Action is also a method of Reinforcement Learning. It is an on-policy method and it is defined as over "state-action pair, rather than just the state" [6]. This follows a policy to take the next value of q instead of Q learning which has a greedy approach and doesn't follow a policy. Expected SARSA takes the mean of all the values of Q in the current action.
It uses the knowledge regarding the stochasticity in policy to perform updates that has a lower variance which in turn leads to better learning rate. Here α is the learning rate.

Update Rule for Expected SARSA

$$Q(s_t, a_t) \leftarrow Q(s_t, a_t) + \alpha[r_{t+1} + \gamma \sum_a \pi(s_{t+1}, a) Q(s_{t+1}, a) - Q(s_t, a_t)]$$

### 3.3 DDQN (Double Deep Q Networks)

The major concern with Deep Q learning is that it is known to overestimate the action values and hence negatively affect the performance. In DQN, the selection and evaluation of action are coupled together whereas, in Double DQN, they are decoupled i.e. two value functions are learned, one for estimating the value of the greedy policy and other for evaluating the value for this policy [5]. 2 layers of CNN are used for this function. Its update is the same as for DQN, except for replacing the weights of the target network. Update Rule:

$$Y_t^{DoubleDQN} \equiv R_{t+1} + \gamma Q(S_{t+1}, \text{argmax} Q(S_{t+1}, a; \theta_t), \theta_t^-)$$

This means that, as in Q-learning, we are still estimating the value of the greedy policy according to the current weights, as defined by $\theta_t$. However, the final weights of the second network, are replaced with the weights of the target network $\theta_t^-$ for the evaluation of the current greedy policy [5].
It has been experimentally shown to reduce the bias in the results.

## 4 Experimental Results

- **Methodology**: In the previous section, we discussed the three algorithms of reinforcement learning that we implemented to develop an AI model for Dino Run i.e. Deep Q Learning, Double Deep Q Learning and Expected SARSA. The main task for this project was to perform experiments and compare results of each algorithm. For this, DQN has been considered as the base for all the experiments. DQN was trained first and the results obtained from it were used to compare the results of other two algorithms. The input files which are fed into the model are the captured images of screen while the game is in playing mode. Four consecutive images are used as a single input which makes its dimension very high. So, these images are transformed into grey scale with its dimension converted to 80*80.

  After convoluting these images further with our CNN model, we start the actual prediction of q-values by taking an initial state (s_t) and action as do nothing. A matrix of 2*1 is obtained as an output which consists of two q-values representing the maximum reward for each action and the next action is selected based on the maximum q-value obtained. This best action is then used to predict the next set of q values for next state. All these values – current state, action, reward, next state and terminal (which is a Boolean variable with value True if the agent died and False otherwise) are stored in a Replay Memory, taken as 50k, which serves as experience obtained by our agent. For Expected SARSA, we store the next action as well in this cache. In DDQN, we use 2 CNN layers to update the weights of networks. Initially, we let the agent observe the environment for OBSERVATION = 1000 number of steps, followed by the exploration process till timestep =100000. After this, the training phase begins. As the game proceeds, we use epsilon-greedy approach, to randomly let the agent decide if it wants to explore or exploit the environment. We use batch learning to make use of agent's experience stored in Replay Memory to train the CNN. It helps to increase the efficiency of the learning. Random batches are selected from this cache for training.

In our experiments, we are calculating Q(s,a) values with CNN and storing them in the cache/Replay Memory. Since, we are saving the immediate returns, and we are using a Markov environment, the latest policy values are just as important as the old policy values and the experience gained by the agent doesn't get old. Evaluation metric used for CNN model training are the loss values which are calculated using mean square error. We also capture scores obtained at the end of each episode. All the hyperparameters used are same for all the experiments including minibatch of size 16, learning rate of 1e-4 and the decay rate i.e. gamma of 0.99. Furthermore, Adam Optimizer is used for determining adaptive learning rates for our CNN model.

The graph plotted to compare performance of each algorithm are scores vs timesteps. Each of the graphs of DQN is compared to graphs of other two algorithms and plotted using Matplotlib library of python.

- **Results**: In our algorithms, an "episode" consists of numerous timesteps. A timestep is a counter value which gets updated each time the agent takes an action in the current state and we predict policy values for the current state-action pair. Since, we are using CNN to predict these values, we also predict the next policy values based on next state and the best action (obtained from taking max of current q values) in the same timestep to satisfy the Bellman equation. An episode ends when the player/agent dies. The score of the game is stored at this point. We use mean squared error between current policy value and optimal policy value to measure loss function for our CNN.

    We ran each of the three algorithms for ~8 hours using only i5 CPUs to compare the performance and scores of each. We have considered DQN as our base algorithm as it was implemented in [1]. We compare the results of Expected SARSA (on-policy) and DDQN (off-policy) against DQN (off-policy). We see in *Table 1* the scores versus absolute number of timesteps for total training time, which is variable for each algorithm. *Figure 3* shows this comparison as a bar chart where scores and timesteps have been normalised to consider training time of 8 hours for each algorithm. DQN obtains score of 2351 in 429400 timesteps, DDQN obtains 2800 score in 260861 timesteps and Expected SARSA gets a score of 405 in 300990 timesteps. *Figure 4* and *Figure 5* show the comparisons of average scores per epochs vs number of epochs. Here an "epoch" is 10 episodes. However, the point to note here is that the length of an episode varies and depends on the state-action pair decided by the agent as it only ends when the agent dies, i.e., does not take an appropriate action and runs into an obstacle. These figures show that although DDQN learns slower than DQN but it learns the best as is evidenced by the highest score achieved in minimum

number of timesteps. They also show that Expected SARSA learns the slowest and performs the worst as compared to the other two. This concludes that Expected SARSA is the slowest to converge and DDQN learns the best with respect to DQN.

|  | SARSA | DQN | DDQN |
|---|---|---|---|
| ***Timestep*** | 300990 | 429400 | 260861 |
| ***Max Score*** | 405 | 2351 | 2800 |
| ***No. of Episodes*** | 4114 | 2295 | 2647 |
| ***Average length of episode*** | 73.16 | 187.10 | 98.54 |

Table 1: Timesteps taken to reach the max score

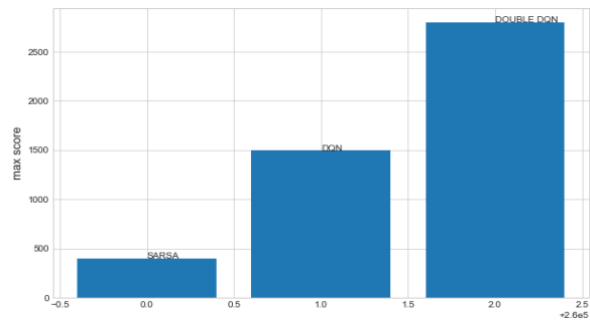

Figure 3 Performance comparison

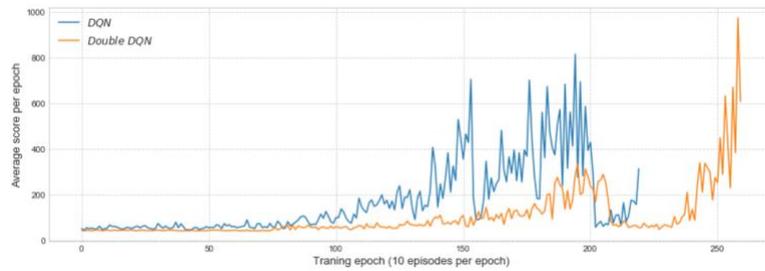

Figure 4 Score Comparison for DQN and DDQN

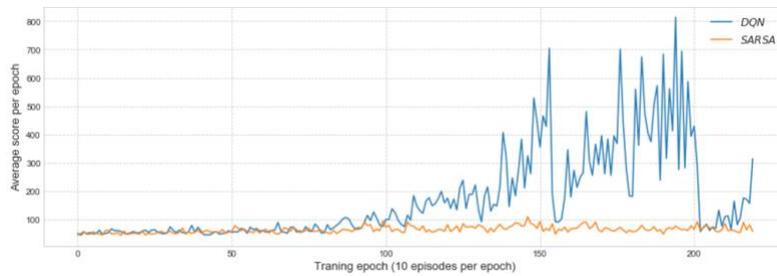

Figure 5 Score Comparison for DQN vs SARSA

- **Discussion**: During the course of this experiment, we realised that we could not use the tabular representation of the Q-values here (as in Q-learning algorithm) because the state-action space is infinitely large and continuous. In this case, we had to transition to use a neural network to predict the value of a state and action pair.

[6] claims that Expected SARSA performs better than Q-Learning and SARSA. However, based on the results discussed above, we see that Expected SARSA algorithm when used with CNN and run for the same duration as DQN and DDQN, takes the longest time to learn and converge. This is because we are taking mean of the policy values instead of taking the maximum as in DQN and DDQN. We also calculate next action along with next state before training CNN and store it in our Replay Memory for it. This is done to honour the major difference between Q-learning and SARSA algorithms (SARSA considers the next state and action value unlike Q-Learning which considers just the next state). We also see that DDQN, although runs slower than DQN, learns the best. This is because it uses 2 layers of CNN to predict the policy values. DQN performs midway through DDQN and Expected SARSA.

We interpret the major difference between Expected SARSA and DQN to be between the way we calculate Q-values. For DQN, we use the maximum Q-value to find the best action for the current state and using this action value predict the maximum Q value of the next state, for every exploration and exploitation step because we assume that the goal of the agent is to get the best estimate of the next Q-value. But, if we know nothing about the environment we are exploring, this introduces maximization bias (noise) in the results for DQN. This causes over-estimation of Q-values. In Expected SARSA, we calculate maximum Q-value for exploration step, to find the best action, but mean Q-values for exploitation step (in training). This reduces the positive bias in Q-values but as a result makes learning slower for it as compared to DQN. On the other hand, DDQN has slower convergence rates (as opposed to DQN) due to use of 2 layers of CNN – 1 for DQN and another for Target network, instead of the 1 layer in DQN but it learns the best and finds better policies since, the maximization bias of DQN has been mitigated here by reducing overestimations of policy values by decoupling the maximization operation in the target network into action selection and action evaluation [5].

## 5   Conclusions

In this paper we studied different temporal difference approaches to train the agents for playing Chrome Dino run and implemented three state-of-the-art algorithms, that is Deep Q-Learning, Double Deep Q-Learning and Expected SARSA. In the given scenario, we observe that off policy algorithms (DQN and DDQN) perform better than the on policy SARSA algorithm. Analysing the results, we also noticed that double deep Q learning learns at a slow rate in the start but gives the best performance after certain number of timesteps. And Expected SARSA learns at a very slow rate and does not show significant score improvements in the given timeframe, while we see a uniform learning and score improvement in case of DQN. Another important deduction from this is the length of an episode is longest in case of DQN, which shows that DQN is the more consistent compared to other two algorithms [*Table 1*].

In the future work, these models can be trained on GPU based system for a longer timeframe, to have a more significant result. Furthermore, some algorithms such as Deep Deterministic Policy Gradient (DDPG), and Duel DQN can be implemented for this game and have an extensive comparison of the stated results. Another important observation is that the speed of obstacles increases with the increase of score, so the game can be divided into stages. Each stage is given an input from the previous stage but are trained individually. This might improve the performance of RL algorithms further and reduce the variance in scores.